\documentclass[letterpaper]{article} 
\usepackage{aaai2026}  
\usepackage{times}  
\usepackage{helvet}  
\usepackage{courier}  
\usepackage[hyphens]{url}  
\usepackage{graphicx} 
\usepackage{natbib}  
\usepackage{caption} 
\usepackage{amsmath}
\usepackage{xcolor}

\usepackage{array}
\newcolumntype{P}[1]{>{\centering\arraybackslash}p{#1}}

\usepackage{multirow}

\usepackage{subcaption}

\usepackage{booktabs}

\usepackage{enumitem}
\newlist{questions}{enumerate}{2}
\setlist[questions,1]{label=\textbf{RQ\arabic*},ref=RQ\arabic*}
\setlist[questions,2]{label=(\alph*),ref=\thequestionsi(\alph*)}

\title{LiRA: A Multi-Agent Framework for Reliable and Readable Literature Review Generation}
\author{
    Gregory Hok Tjoan Go\textsuperscript{\rm 1, 2},
    Khang Ly\textsuperscript{\rm 2},
    Anders Søgaard\textsuperscript{\rm 3 },\\
    Seyed Amin Tabatabaei\textsuperscript{\rm 2},
    Maarten de Rijke\textsuperscript{\rm 1},
    Xinyi Chen\textsuperscript{\rm 1}
   }
\affiliations{
    \textsuperscript{\rm 1}University of Amsterdam\\ 
    \textsuperscript{\rm 2}Elsevier B.V.\\ 
    \textsuperscript{\rm 3}University of Copenhagen\\
    
    \{g.go, k.ly, s.tabatabaei\}@elsevier.com, soegaard@di.ku.dk, 
    \{m.derijke, x.chen2\}@uva.nl
}

\begin{document}

\maketitle

\begin{abstract}
The rapid growth of scientific publications has made it increasingly difficult to keep literature reviews comprehensive and up-to-date. Though prior work has focused on automating retrieval and screening, the writing phase of systematic reviews remains largely under-explored, especially with regard to readability and factual accuracy. To address this, we present \textbf{LiRA} (\textbf{Li}terature \textbf{R}eview \textbf{A}gents), a multi-agent collaborative workflow which emulates the human literature review process. LiRA utilizes specialized agents for content outlining, subsection writing, editing, and reviewing, producing cohesive and comprehensive review articles. Evaluated on SciReviewGen and a proprietary ScienceDirect dataset, LiRA outperforms current baselines such as AutoSurvey and MASS-Survey in writing and citation quality, while maintaining competitive similarity to human-written reviews. We further evaluate LiRA in real-world scenarios using document retrieval and assess its robustness to reviewer model variation. Our findings highlight the potential of agentic LLM workflows, even without domain-specific tuning, to improve the reliability and usability of automated scientific writing.
\end{abstract}

\begin{links}
    \link{Code}{www.github.com/lira-workflow/auto-review-writing}
    \link{Published version}{https://ojs.aaai.org/index.php/AAAI/article/view/41489}
\end{links}

\section{Introduction}

Since their inception, literature reviews have been consistently used to streamline the advancement of various scientific fields \cite{snyder_identifying_2016}. Of these reviews, one of the most important types is the Systematic Literature Review (SLR), which reproducibly synthesizes a significant portion of existing research relating to a specific research question being addressed \cite{kitchenham_guidelines_2007, bangdiwala_importance_2024}. This role has become increasingly more crucial, evidenced by how quite a few researchers consider them to be original research or potentially even a mandatory step in the scientific process itself \cite{kraus_importance_2023, palmatier_review_2018}.

Due to the large amount of new findings and research being disseminated through publications, it has become very difficult to release SLRs in a timely fashion \cite{qi_generation_2025, ofori-boateng_towards_2024, tian_overview_2025}. For example, the estimated time required to complete an SLR has increased significantly in the past few decades in the medical domain \cite{allen_estimating_1999, borah_analysis_2017}, which is further compounded by the necessity of using expert labor \cite{atkinson_ai-pocalypse_2025}. In tackling this, the vast majority of research relating to SLR automation focuses on the retrieval and screening of scientific papers \cite{orel_automated_2023, marshall_toward_2019, chen_can_2025}, as these are the most time-consuming steps \cite{chai_research_2021}. However, there remains the task of compiling the findings into a comprehensive review paper. Only a small number of works have been published \cite{kasanishi_scireviewgen_2023, qi_generation_2025, wang_autosurvey_2024}, let alone those which focus on the readability and hallucination mitigation aspects.

In this work, we present \textbf{LiRA} (\textbf{Li}terature \textbf{R}eview \textbf{A}gents), an agentic solution aimed at addressing the minimal research related to automated literature review writing. It is a Large Language Model (LLM)-based agentic workflow building upon existing relevant works and integrates some of the most recent SLR-writing guidelines to generate accurate and high-quality reviews, with an additional emphasis on readability. Moreover, it is entirely out-of-the-box, requiring no task-specific pre-training or fine-tuning. We also reduce hallucination in the outputs, which is one of the main barriers in trustworthy automated writing and a key factor preventing the widespread use of similar Artificial Intelligence (AI)-powered systems \cite{alkaissi_artificial_2023, oconnor_large_2024, xu_critical_2023}.

To demonstrate LiRA's capabilities, we propose the below research questions:
\begin{questions}[leftmargin=*]
    \item \textit{To what extent are the qualitative Systematic Literature Reviews created by LiRA similar to human-written ones compared to existing literature writing methods when all are given the same set of references?}
    \item \textit{How well written are the generated articles compared to existing literature writing methods when all are given the same set of references?}
    \item \textit{How well can LiRA properly use citations from the provided sources to generate qualitative Systematic Literature Reviews compared to other methods?}
    \item \textit{How well can LiRA be used in real-world settings when using references returned by a scientific document retriever?}
\end{questions}

\noindent
We summarize our main contributions as follows:
\begin{itemize}
    \item To the best of our knowledge, we propose the first agentic LLM-based automated literature review writing workflow which emulates the human writing process and integrates the findings of other relevant agentic workflows.
    \item We explore the usage of formally defined guidelines and techniques from relevant similar fields in the agentic workflow, such as the idea of thoroughly analyzing the existing papers before beginning the writing process, establishing a crucial link between theory and application.
    \item We establish several state-of-the-art baseline results for the automated literature review writing task across multiple settings, comparing between existing systems when using the same LLM type throughout.
\end{itemize}

\section{Related Work}

\paragraph{Agentic workflows} Comparisons have been made between agentic LLM systems and human cognition, due to how breaking down a task into smaller steps, which these workflows often do, is commonly used to describe how humans solve more intricate problems \cite{flower_cognitive_1981, correa_humans_2023}. Using this concept, several works show promise in implementing agentic workflows in various fields, such as the medical sciences \cite{tang_medagents_2024} and law \cite{watson_law_2025}. In some cases, this results in an improvement of more than 90\% compared to a simple baseline \cite{watson_law_2025}. 

However, for the task of automated writing specifically, only few works have been published thus far~\cite{qi_generation_2025, wang_autosurvey_2024, shao_assisting_2024, tian_overview_2025}. Of these works, only two of them address literature reviews or a similar document type and include open-source code \cite{qi_generation_2025, wang_autosurvey_2024}. Neither paper takes output lengths and how they relate to the readability and evaluation of each proposed system using the listed metrics into account. As a result, no works seem to exist which tackle the issues of readability, and only minimal work exists in addressing the factuality of the resulting articles.

\paragraph{Literature review} For centuries \cite{lind_treatise_2014}, the process of writing literature reviews has been considered crucial for the development of science \cite{meerpohl_scientific_2012, higgins_cochrane_2011, chalmers_brief_2002}, as it provides researchers valuable insight on which research questions to answer via an analysis of earlier works \cite{chalmers_avoidable_2009, eagly_using_1994}. Moreover, improvements have been made to eliminate personal biases \cite{egger_introduction_2001} through the introduction of the SLR, which uses systematic methods of review for the collation and synthesis of findings \cite{randles_guidelines_2023, snyder_literature_2019, page_prisma_2021}. 

Given how time-consuming \cite{borah_analysis_2017} and costly \cite{michelson_significant_2019} this process is, a need for a viable alternative has emerged. Despite this, there is not much relevant innovation in natural language processing that tackles this issue \cite{mohammad_using_2009, kasanishi_scireviewgen_2023, agarwal_towards_2011}, let alone results indicating real-world usability. Therefore, we address all aforementioned problems by introducing an agentic workflow which uses LLMs to generate literature reviews automatically, demonstrating its potential by focusing on both readability and factuality.

\section{The LiRA Framework}

LiRA emulates the human literature review process by decomposing it into specialized, interacting LLM-based agents. Each agent tackles a key sub-task: either structural planning, fine-grained writing, consistency refinement, or factual verification, which results in a modular and scalable pipeline. This section introduces the core agents and their design motivations.

\begin{figure}[!t]
    \centering
    \includegraphics[width=\linewidth]{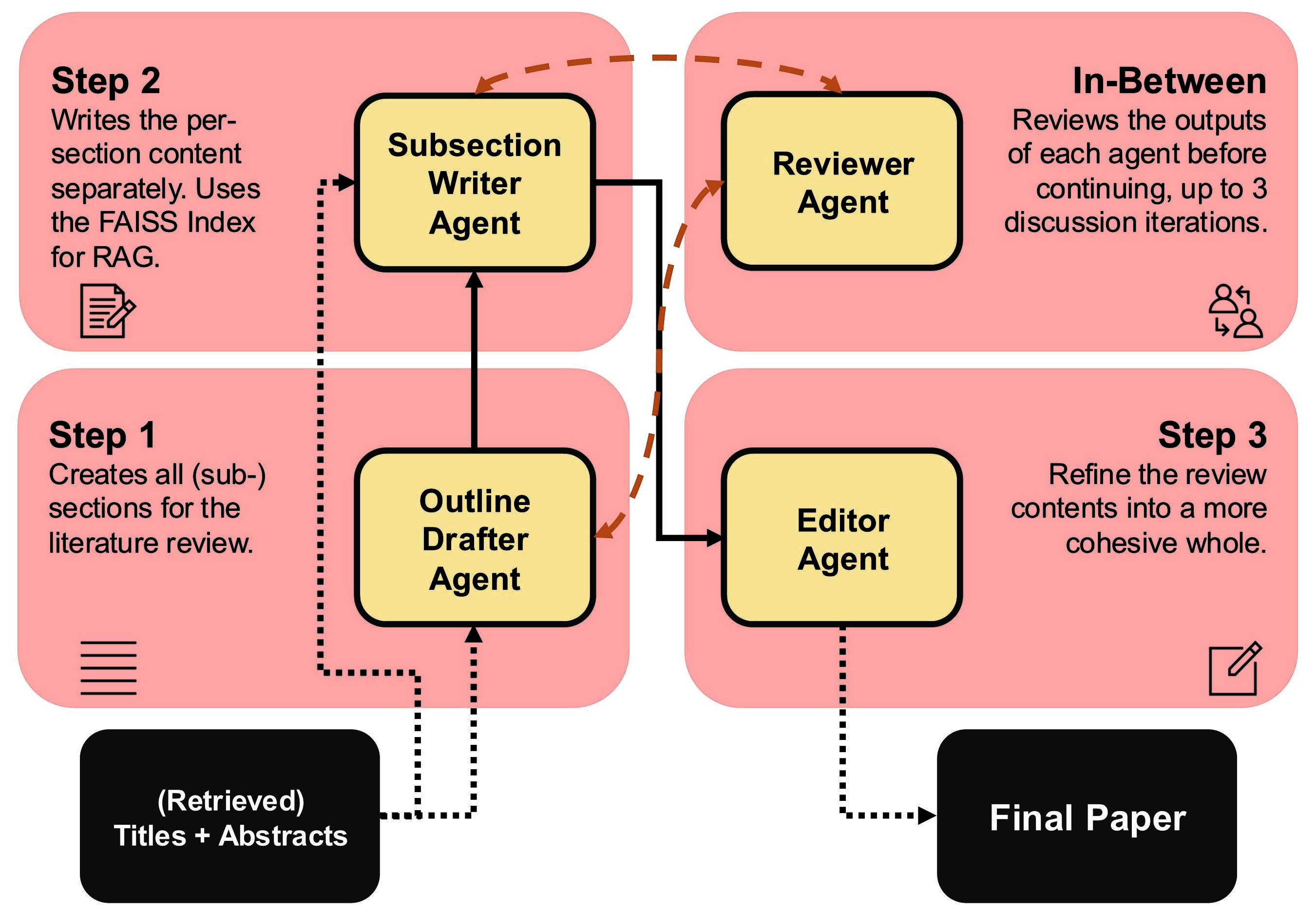}
    \caption{An overview of the LiRA architecture. The narrow dotted arrows represent document input/output, the wide dotted arrows indicate the refinement process, and the filled-in arrows signify the system's main flow. Each agent is explained in the below sections.}
\end{figure}

\subsection{Outline Drafter Agent}

A key challenge in literature review writing is constructing a coherent structure from a large and unorganized set of references. Instead of relying on the model to implicitly determine the structure during generation, LiRA explicitly drafts an outline to guide the writing process. This agent takes the topic and abstracts (or alternatively the full texts) of the provided references to produce a set of candidate outlines consisting of main sections and subsections. These are then combined into a unified draft structure that includes descriptions for each section and suggested supporting papers.

To manage the context size and focus on only the
most relevant content, the outline is constructed from up to 50 references. We use existing heuristics \citep{wang_autosurvey_2024} that recommend generating approximately 8 sections with around 4 subsections, but the agent can adapt this amount as needed. We also include default sections such as an introduction and conclusion to ensure consistency and completeness.

\subsection{Subsection Writer Agent}

Generating a full review in a single pass risks superficial coverage and poor section coherence. We tackle this by having a subsection writer agent, which writes each (sub)section individually in parallel, conditioned on the description and a selected subset of relevant references. This design encourages fine-grained generation and makes the outputs more modular and easier to revise.

Relevant references are retrieved using the FAISS index \citep{douze_faiss_2025} based on section-level descriptions, capped at 25\% of the total pool per subsection (min 3, max 150). This balances citation diversity with input tractability. Moreover, the reference titles, abstracts, or full texts can be retrieved depending on the availability. The agent outputs $\sim$1,000 words per subsection, enabling long-form synthesis while staying within model context limits. Also, the article title, abstract, and conclusion are written after the body is generated, mimicking human writing practices and avoiding the premature commitment of top-down generation.

\subsection{Editor Agent}

Even with structural planning and localized generation, the assembled review may exhibit issues such as redundancy, inconsistency, or stylistic mismatches across sections. These are addressed by an editor agent, which refines the entire draft with a focus on presentation and style. It performs standard editing operations, including improving transitions, enhancing vocabulary, and ensuring overall fluency and cohesion. Importantly, this agent does not alter the factual content, preserving the integrity of the generated information.

If the generated output exceeds the model’s context window, the agent detects whether the text ends abruptly—typically when the maximum output length (16,384 tokens) is reached. In such cases, the model is prompted to continue the output using the original input and previously edited content as context. This mechanism effectively doubles the generation capacity, though at the cost of additional input overhead.

\subsection{Reviewer Agent}

To improve factual accuracy and review quality, we introduce an LLM-based reviewer agent inspired by human editorial workflows. This agent evaluates intermediate outputs (e.g., outlines) based on adapted criteria from systematic review guidelines \citep{snyder_literature_2019}, including content completeness, transparency, clarity, and contribution.

If a component (e.g., an outline or section) fails to meet quality thresholds, the reviewer provides structured feedback and triggers regeneration. The review loop continues up to 3 rounds before fallback progression, balancing refinement with computational efficiency.

\subsection{Citation Behavior}

Citation hallucination remains a major concern in automated scientific writing, as models may generate plausible-sounding but non-existent references. To address this, LiRA incorporates citation grounding directly into the generation process. Rather than relying on abstract placeholders (e.g., numbered citations), agents cite sources using full article titles, which act as semantic anchors and help the model maintain alignment with the provided references.

After generation, these in-text citations are post-processed into standard numbered references, and any hallucinated titles are redacted during evaluation to ensure fair comparison. This approach improves factual consistency and ensure fair comparison during evaluation.

\subsection{Additional Implementation Details}

All agents in LiRA were implemented using LangGraph. Moreover, each agent has its own memory, a standard practice for LLM-based agentic workflows to better act upon feedback \cite{sumers_cognitive_2023, qian_chatdev_2024}. This system of storing feedback in memory is comparable to Reflexion  \cite{shinn_reflexion_2023}, where the agent has to adjust its behavior based on the feedback provided. Parallelization is also included to increase the processing speed of certain steps, namely the researcher when analyzing papers and the content writer for generating the article (sub)sections.

Aspects from the design of MetaGPT \cite{hong_metagpt_2023} were adapted for more efficient inter-agent communication. Specifically, all agents are required to return their outputs as structured documents to avoid potential inefficiencies relating to information presentation, which are then sent to unique shared message pools for quick information retrieval. Moreover, the input contents are filtered based on the model's maximum context window length (128,000 tokens in our case) to prevent information overload. As a method of improving the model's output quality with minimal intervention, Zero-Shot Chain-of-Thought \cite{kojima_large_2022} is included in the prompts for all agents except the researcher and editor, as they do not perform refinement.

\section{Experiments}

\subsection{Baselines}

To evaluate LiRA, we compare it against direct prompting and, to the best of our knowledge, the only two publicly available agentic frameworks for survey writing. For fair comparison, all systems (including LiRA) are implemented with \verb|gpt-4o-mini| as the underlying LLM.

\paragraph{Direct prompting (DP)} As the simplest baseline, we directly prompt the LLM with task instructions and the full set of reference titles and metadata, asking it to generate a review with the specified sections and length. When the reference list exceeds the context window, it is passed as an attached file, requiring the model’s file-reading capability. This baseline tests whether a single prompt can produce a coherent review without decomposition or refinement.

\paragraph{MASS-Survey (MASS)} Introduced in an automated survey-writing challenge \cite{tian_overview_2025}, MASS is the only agentic framework from that challenge with publicly available code. Its workflow differs fundamentally from LiRA: Instead of decomposing the writing process into specialized roles with iterative feedback, MASS follows a strictly sequential pipeline. The system first clusters references by topical similarity to construct an outline, then generates section contents and a title directly from these clusters, and finally appends a conclusion. Long reference lists are handled by passing them as attachments.

\paragraph{AutoSurvey (AS)} AutoSurvey \cite{wang_autosurvey_2024} is a multi-stage framework for automatically generating literature surveys in computer science. Given a query (in our case, the original review title), it retrieves relevant publications and uses their titles and abstracts to construct an outline, followed by subsection drafting with refinement steps and partial use of retrieved article content (up to 1,500 tokens). While the paper claims retrieval-grounded drafting, we did not find corresponding functionality in the released code.

For fair comparison, we modified AutoSurvey to restrict retrieval to the references cited in the target human-written review. This required reducing the number of candidate documents per subsection to between 2 and 25\% of the references, or falling back to 60 when the fraction exceeded this threshold (as in the original design). We also generalized system prompts from “\textit{You are an expert in Artificial Intelligence}” to “\textit{You are an expert in a relevant field}” to make the system better able to generate articles for multiple domains.

\subsection{Metrics}
To comprehensively evaluate the generated literature reviews, we consider three complementary dimensions: content similarity to human-written reviews, writing quality, and citation reliability.

\paragraph{Similarity to the human-written review} We measure how similar the generated literature reviews are to human-written ones. Metrics include ROUGE-L, heading soft recall (\textit{hsr}), heading entity recall (\textit{her}), and article entity recall (\textit{aer}). Together, these capture lexical overlap, structural alignment, and coverage of key cited works. Full definitions are provided in Appendix~\ref{appendix:sim_metrics}.

\paragraph{Writing quality} We evaluate writing quality using both automatic and human assessments. For automatic evaluation, we use the Prometheus 2 LLM \cite{kim_prometheus_2024}. It is an open-source LLM evaluator that uses the appropriate reference materials (the instruction, reference answer, and score rubric) to provide assessments which mostly align with those of human annotators. Three aspects are evaluated for all generated articles, namely the coverage, structure, and relevance. Here, coverage represents how broad the subject matter of the review is, while structure measures the organization and flow of the review, and relevance indicates how well the review is able to stay on-topic overall. Additional specifications for the model and how it measures the aforementioned aspects can be found in Appendix~\ref{appendix:llme}.

For human evaluation, we employed subject matter experts (SMEs) who helped evaluate the outputs of the system on the same aspects as mentioned above. For feasibility, the annotation was performed differently for each dataset, though the title, outline, and a section snippet were utilized in both cases. For SciReviewGen, we employed a group of 3 SMEs from Straive to select their preferences between human- and LiRA-written articles for 30 sample snippets, using a rubric as guidance for determining their choice. It must also be noted that the order in which these samples were presented was randomized for each row.

Meanwhile, a dedicated team from within the company provided scores ranging from 1 to 5 for the AutoSurvey and LiRA articles while using the human-written ones as a baseline. This grading was done using the same rubric as mentioned above. Furthermore, due to this type of annotation being more labor-intensive, only 15 article snippets were used.

\paragraph{Citation quality} We evaluate how well generated claims are grounded in appropriate references. Our metric, Citation Quality F1-Score (CQF1), balances precision (penalizing irrelevant or hallucinated citations) and recall (capturing missing but necessary citations), serving as a proxy for hallucination in scientific writing. A full overview of the details is given in Appendix~\ref{appendix:citation_q}.

\subsection{Datasets}

We evaluate the effectiveness of the LiRA framework using two datasets. The primary dataset is SciReviewGen \cite{kasanishi_scireviewgen_2023}, a publicly available benchmark built on the Semantic Scholar Open Research Corpus (S2ORC) \cite{lo_s2orc_2020}. It contains 10,000 review articles in computer science, referencing approximately 690,000 papers. Each review is annotated with structured metadata, including titles, section headers, full texts, and references. Following the setup in \citet{shao_assisting_2024}, we randomly sample 125 reviews, ensuring each selected article contains a sufficient number of references—averaging around 70 per paper.

To assess the generalizability beyond computer science, we additionally evaluate on an internal dataset of 125 expert-written reviews from ScienceDirect, covering 23 subject areas including business, microbiology, and materials science. The dataset is matched in size to the SciReviewGen subset to support direct cross-domain comparison.

\subsection{Results}

Across both datasets, LiRA consistently outperforms baseline systems on the majority of evaluation metrics, demonstrating stronger alignment with human-written reviews, higher writing quality, and more reliable citation use.

\paragraph{Similarity to human-written review} LiRA achieves the highest ROUGE scores, indicating stronger lexical alignment with human-written reviews. AutoSurvey attains slightly higher heading/entity recall, but largely due to verbosity: on average, AutoSurvey produces 50,000 tokens per article compared to only 22,000 for LiRA. Since recall-based metrics do not normalize for length, longer outputs are naturally favored. Crucially, this shows that LiRA generates concise yet information-dense reviews, rather than inflating scores by producing excessive text.

\paragraph{Writing quality} LiRA achieves the best overall writing quality, with a clear advantage in structural coherence. AutoSurvey performs marginally better in coverage, again reflecting its longer outputs, but at the cost of organization and readability. SME evaluations align with these trends: on ScienceDirect, experts strongly preferred LiRA for structure, while AutoSurvey received marginally higher scores for coverage and relevance, suggesting a trade-off between breadth and coherence. On SciReviewGen, SMEs favored LiRA over the human-written reviews, noting its broader and more balanced coverage given the outlines. These results highlight LiRA’s strength in generating well-structured, concise, and expert-aligned reviews, in some cases even being favored over human-written baselines.

\paragraph{Citation quality} LiRA demonstrates the largest gains in citation reliability, achieving the highest Citation Quality F1 (CQF1) scores across both datasets (0.76 on SciReviewGen, 0.73 on ScienceDirect) and substantially outperforming AutoSurvey ($\leq$0.63) and all other baselines. This indicates that LiRA is more effective at grounding claims in appropriate references, avoiding both omissions (recall errors) and hallucinations (precision errors), which stem from LiRA’s citation-grounded generation design that explicitly enforces reference anchoring during drafting and refinement.

From this, it can be seen that LiRA overall produces literature reviews that are concise, structurally coherent, and citation-faithful, while maintaining competitive coverage. This balance between quality and reliability highlights LiRA as a more trustworthy and practically useful framework for automated survey writing. achieves the highest ROUGE scores, indicating stronger lexical alignment with human-written reviews. AutoSurvey attains slightly higher heading/entity recall, but largely due to verbosity: on average, AutoSurvey produces 50,000 tokens per article compared to only 22,000 for LiRA. Since recall-based metrics do not normalize for length, longer outputs are naturally favored. Crucially, this shows that LiRA generates concise yet information-dense reviews, rather than inflating scores by producing excessive text.

\begin{table}[t!]
\centering
\begin{tabular}{@{}lcccc@{}}
\toprule

\multicolumn{1}{c}{\textbf{Metric}} & \textbf{DP} & \textbf{MASS} & \textbf{AS}         & \textbf{LiRA}        \\ 
\midrule
\multicolumn{5}{c}{\textbf{SciReviewGen}}                                                                      \\ \midrule
ROUGE                               & 0.06 ± 0.0  & 0.09 ± 0.0    & 0.09 ± 0.0          & 0.13 ± 0.0           \\
\textit{hsr}                        & 0.69 ± 0.1  & 0.66 ± 0.1    & \textbf{0.92 ± 0.4} & 0.82 ± 0.1           \\
\textit{her}                        & 0.06 ± 0.0  & 0.05 ± 0.0    & \textbf{0.15 ± 0.0} & 0.10 ± 0.0           \\
\textit{aer}                        & 0.06 ± 0.0  & 0.09 ± 0.0    & \textbf{0.34 ± 0.0} & 0.27 ± 0.0           \\ \midrule
\multicolumn{5}{c}{\textbf{ScienceDirect}}                                                                     \\ \midrule
ROUGE                               & 0.02 ± 0.0  & 0.04 ± 0.0    & \textbf{0.13 ± 0.0} & \textbf{0.13 ± 0.0}  \\
\textit{hsr}                        & 0.24 ± 0.2  & 0.22 ± 0.2    & 0.24 ± 0.4          & \textbf{0.25 ± 0.1}  \\
\textit{her}                        & 0.03 ± 0.0  & 0.03 ± 0.0    & \textbf{0.13 ± 0.0} & 0.05 ± 0.0           \\
\textit{aer}                        & 0.03 ± 0.0  & 0.05 ± 0.0    & \textbf{0.25 ± 0.0} & 0.17 ± 0.0           \\ \bottomrule
\end{tabular}
\caption{Results for similarity to the human-written reviews with the baseline settings.}
\label{results:similarity}
\end{table}

\begin{table}[t!]
\centering
\begin{tabular}{@{}lcccc@{}}
\toprule

\multicolumn{1}{c}{\textbf{Metric}} & \textbf{DP} & \textbf{MASS}       & \textbf{AS}         & \textbf{LiRA}       \\ \midrule
\multicolumn{5}{c}{\textbf{SciReviewGen}}                                                                           \\ \midrule
Coverage                            & 3.53 ± 0.4  & 4.30 ± 0.3          & \textbf{4.50 ± 0.1} & 4.45 ± 0.1          \\
Structure                           & 3.15 ± 0.9  & 2.47 ± 1.2          & 2.30 ± 1.3          & \textbf{3.38 ± 0.9} \\
Relevance                           & 4.49 ± 0.2  & \textbf{4.74 ± 0.2} & 4.55 ± 0.2          & 4.57 ± 0.2          \\
Average                             & 3.72        & 3.83                & 3.78                & \textbf{4.13}       \\ \midrule
\multicolumn{5}{c}{\textbf{ScienceDirect}}                                                                          \\ \midrule
Coverage                            & 3.08 ± 1.0  & 3.44 ± 1.0          & \textbf{4.10 ± 0.1} & 3.90 ± 0.3          \\
Structure                           & 3.23 ± 1.0  & 2.59 ± 1.1          & 2.21 ± 1.3          & \textbf{3.42 ± 0.9} \\
Relevance                           & 3.98 ± 0.7  & 4.15 ± 0.7          & 4.29 ± 0.3          & \textbf{4.33 ± 0.3} \\
Average                             & 3.43        & 3.39                & 3.53                & \textbf{3.88}       \\ \bottomrule
\end{tabular}
\caption{Writing quality results for the baseline settings.}
\label{results:quality}
\end{table}

\begin{table}[t!]
\centering
\begin{tabular}{@{}lcccc@{}}
\toprule
\textbf{Dataset} & \textbf{DP} & \textbf{MASS} & \textbf{AS} & \textbf{LiRA} \\ \midrule
SciReviewGen     & 0.14        & 0.13          & 0.63        & \textbf{0.76} \\
ScienceDirect    & 0.06        & 0.33          & 0.55        & \textbf{0.73} \\ \bottomrule
\end{tabular}
\caption{Citation quality results for the baseline settings.}
\label{results:citation}
\end{table}

\begin{figure}[t!]
    \centering
    \includegraphics[width=\linewidth]{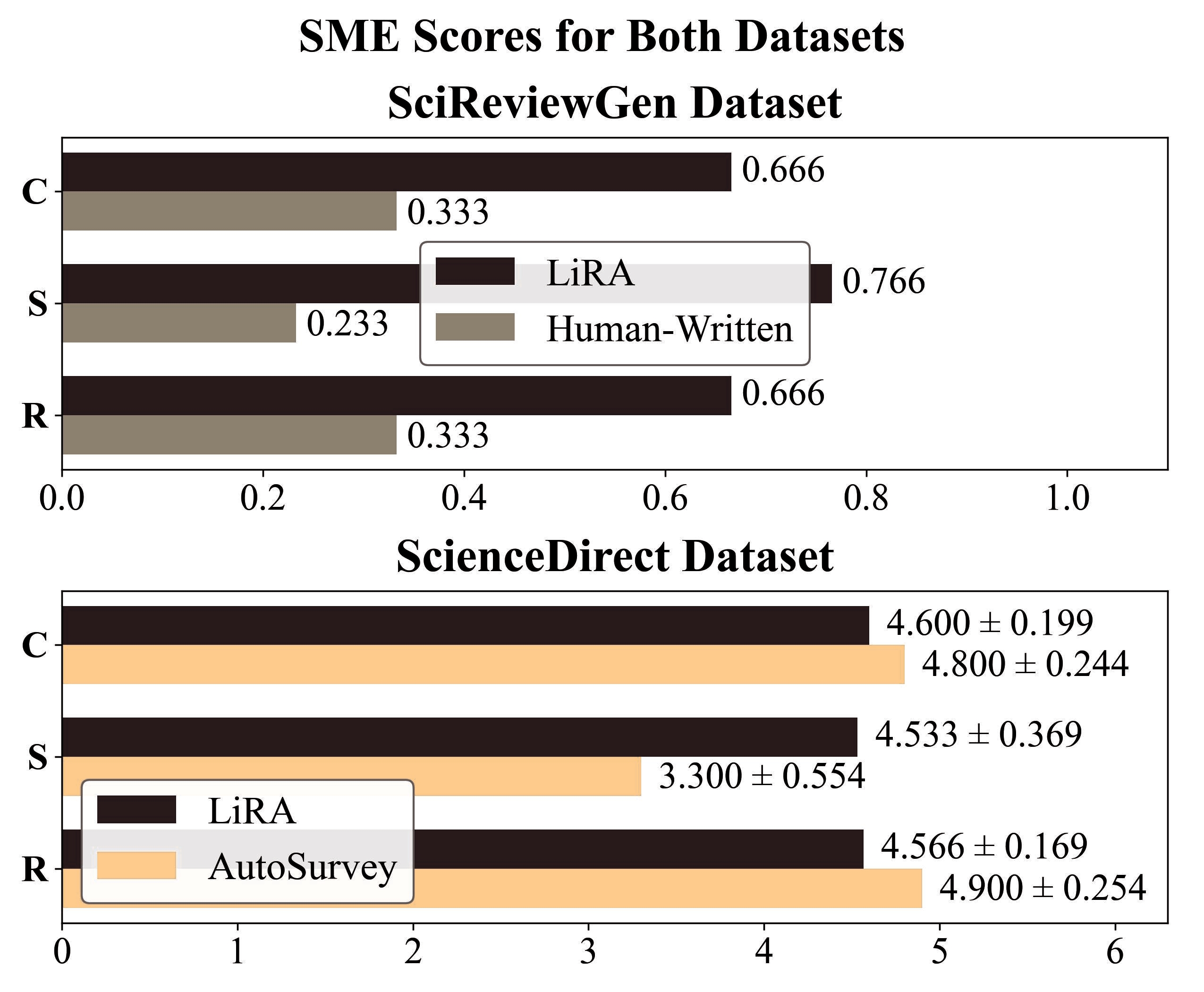}
    \caption{SME evaluation results. Here, \textbf{C} indicates Coverage, \textbf{S} indicates Structure, and \textbf{R} indicates Relevance.}
    \label{results:sme}
\end{figure}

\section{Potential Modifications on LiRA}

In this section, we discuss the adjustments tested on LiRA to evaluate its performance more robustly. This involves the usage of a different LLM type for the reviewer agent to potentially mitigate self-bias amplification in the refinement process, and document retriever usage to evaluate if LiRA can be deployed in real-world settings.

\subsection{Using a different reviewer model}

\paragraph{Method} Based on concerns stemming from self-bias amplification \cite{xu_pride_2024}, experimentation was performed using a reviewer model type different from the one used during generation for each component, which is based on existing suggestions. To this end, the \verb|gemma3:4b| model was used through \verb|ollama| \cite{team_gemma_2025}. This model was chosen because it is open-source, has a context window similar to \verb|gpt-4o-mini|, and has lower hardware requirements compared to most other models. In addition, \verb|ollama| was the selected model provider because of its accessibility and ease of use, with LangGraph already being compatible with it. The parameter values used by \verb|gemma3:4b| were the default \verb|ollama| ones aside from the context window size (128,000) and seed (42).

\paragraph{Results} The results for this setting can be found in Table \ref{results:gemma}. Though the extent of bias mitigation itself cannot be measured properly with the current metrics, we can remark how using \verb|gemma3:4b| has little impact on the scores overall across all metrics. This suggests that alternative model configurations may not significantly alter the article output quality, therefore indicating potentially that the pipeline can work well even when using different LLMs in the process.

\begin{table}[t!]
\centering
\begin{tabular}{@{}lcccc@{}}
\toprule
& \multicolumn{2}{c}{\textbf{SciReviewGen}}                                            & \multicolumn{2}{c}{\textbf{ScienceDirect}}                                          \\
\textbf{Metric} & \textbf{LiRA}       & \textbf{\begin{tabular}[c]{@{}c@{}}LiRA\\ +gemma3\end{tabular}} & \textbf{LiRA}       & \textbf{\begin{tabular}[c]{@{}c@{}}LiRA\\ +gemma3\end{tabular}} \\ \midrule
ROUGE           & \textbf{0.13 ± 0.0} & \textbf{0.13 ± 0.0}                                             & \textbf{0.13 ± 0.0} & \textbf{0.13 ± 0.0}                                             \\ \midrule
\textit{hsr}    & 0.82 ± 0.1          & \textbf{0.83 ± 0.1}                                             & 0.25 ± 0.1          & \textbf{0.25 ± 0.1}                                             \\
\textit{her}    & 0.10 ± 0.0          & \textbf{0.10 ± 0.0}                                             & \textbf{0.05 ± 0.0} & 0.05 ± 0.0                                                      \\
\textit{aer}    & \textbf{0.27 ± 0.0} & 0.26 ± 0.0                                                      & \textbf{0.17 ± 0.0} & 0.17 ± 0.0                                                      \\ \midrule
CQF1            & 0.76                & \textbf{0.77}                                                   & \textbf{0.73}       & 0.73                                                            \\ \midrule
Coverage        & \textbf{4.45 ± 0.1} & 4.45 ± 0.1                                                      & \textbf{3.90 ± 0.3} & 3.89 ± 0.4                                                      \\
Structure       & 3.38 ± 0.9          & \textbf{3.47 ± 0.9}                                             & \textbf{3.42 ± 0.9} & 3.26 ± 1.0                                                      \\
Relevance       & \textbf{4.57 ± 0.2} & 4.64 ± 0.2                                                       & \textbf{4.33 ± 0.3} & 4.29 ± 0.3                                                      \\ \bottomrule
\end{tabular}
\caption{The gemma3 results for the ScienceDirect dataset. Note that if both numbers in a row are bolded, it means they returned the exact same value.}
\label{results:gemma}
\end{table}

\subsection{Retrieval Usage}

\paragraph{Method} All prior experiments assumed the availability of gold references from a reference article to generate a review. This is not the case for real-world settings, however, as novel literature reviews are required to keep up with current developments. Therefore, we evaluate if the system can generate reviews similar enough to the human-written ones when provided with retrieved references instead, hence the inclusion of \textbf{RQ4}. Specifically, an internal API for embedding-similarity search was used, which can be called to retrieve as many references as listed in the human-written review.

\paragraph{Results} We examined if the results when using retrieval differed significantly compared to the baseline researcher setting, which was tested using the appropriate statistical tests. From the results (shown in Table \ref{results:ret}), we note that only two results were significantly different compared to the baseline, indicating that LiRA is capable of performing similarly despite the different references used. More details on the statistical tests can be found in Appendix \ref{appendix:stats}.

\begin{table}[t!]
\centering
\begin{tabular}{@{}lcc@{}}
\toprule
\textbf{Metric} & \textbf{LiRA}          & \textbf{\begin{tabular}[c]{@{}c@{}}LiRA\\ +retriever\end{tabular}} \\ \midrule
ROUGE           & \textbf{0.130 ± 0.021} & 0.128 ± 0.019                                                      \\ \midrule
\textit{hsr}    & \textbf{0.257 ± 0.130} & 0.251 ± 0.116                                                      \\
\textit{her}    & \textbf{0.056 ± 0.043} & 0.054 ± 0.044                                                      \\
\textit{aer}    & \textbf{0.170 ± 0.057} & 0.152 ± 0.054*                                                     \\ \midrule
Coverage        & \textbf{3.892 ± 0.407} & 3.839 ± 0.415*                                                     \\
Structure       & 3.264 ± 1.049          & \textbf{3.411 ± 1.023}                                             \\
Relevance       & \textbf{4.296 ± 0.389} & 4.270 ± 0.372                                                      \\ \bottomrule
\end{tabular}
\caption{The ScienceDirect retrieval results. The stars indicate results significantly lower than the baseline.}
\label{results:ret}
\end{table}

\section{Deployment}

The deployment of LiRA will use the following steps. First, it will be developed using the Python version of LangGraph, which is an open-source and production-ready agentic framework. Furthermore, the \verb|gpt-4o-mini| LLM from AzureOpenAI will be used, with the possibility of using other models given LangGraph's extensive support for various other endpoints such as form \verb|ollama|.

As the use case of LiRA requires it to generate novel literature reviews not based on existing reviews, a document retrieval system will be added on to the system. It would function by asking the user for a review topic as input, which would then be enriched using an LLM (i.e., \verb|gpt-4o-mini|) and afterwards used for embedding-based retrieval using an internal API. This API by default has access to a large collection of scientific articles, and can be replaced depending on the specific circumstances.

\section{Conclusion, Limitations, and Future Work}

This work introduces \textbf{LiRA}, an agentic workflow designed for the automatic writing of literature reviews, which integrates the concepts of research before writing and refinement in its core pipeline. The results obtained show that LiRA is capable of performing the task of automated literature review quite well, outperforming all tested open-source methods when accounting for the varying output lengths, indicating a positive result for essentially every research question proposed. Moreover, it reduces hallucination through improved citation behavior and can demonstrably be used in real-world settings. 

Several improvements could be made, mainly regarding the irreproducibility of results due to the usage of \verb|gpt-4o-mini| for all experiments. This can be solved by using seedable models instead, which should be feasible given current LLM availability. In addition, there is a lack of open-source datasets for this task specifically, which hinders the generalizability of all results to other scientific fields. Therefore, we encourage authors to create additional datasets, ideally in a similar format to SciReviewGen, to facilitate the evaluation of similar systems in the future.

Furthermore, opportunities exist to create more end-to-end pipelines, as the current project does not take into account factors such as primary studies and risk of bias in randomized trials (i.e., the implementation of automated tools based on \citet{cochrane_cochrane_2024}). Doing this would allow for the integration of more steps within the literature review writing process, namely the screening and search criteria definition steps, which would allow for better paper reproducibility.

\section*{Acknowledgments}

We would like to thank Marcela Haldan and Alexandra Noti for their help in providing annotations for the ScienceDirect articles. In addition, this research was (partially) supported by the Dutch Research Council (NWO), under project numbers 024.\-004.\-022, NWA.1389.\-20.\-183, and KICH3.\-LTP.\-20.\-006, and the European Union under grant agreements No. 101070212 (FINDHR) and No. 101201510 (UNITE). Views and opinions expressed are those of the author(s) only and do not necessarily reflect those of their respective employers, funders and/or granting authorities.
\bibliography{aaai2026}

\appendix

\section{Metrics Measuring the Similarity with Human-Written Articles}
\label{appendix:sim_metrics}

This section includes an in-depth explanation of how the textual similarity metrics were implemented and how they function. 
\begin{itemize}
    \item \textbf{ROUGE-L}: This metric is frequently used for text evaluation and functions by counting the number of recalled units in a text from the reference \cite{lin_rouge_2004}. For this project, we used the ROUGE-L metric from the HuggingFace \verb|evaluate| package.
    
    \item \textbf{Heading Soft Recall:} Inspired by the idea of soft recall \cite{franti_soft_2023}, this metric quantitatively measures the heading coverage of a review without relying on exact matches, where a higher score indicates a more comprehensive outline by virtue of being similar to the human-written one. Here, we use the original implementation provided by \citet{shao_assisting_2024}.

    Let $S$ be a set, and let both $P$ and $G$ be sets of predicted and gold headers, respectively. For each item in $S$, the soft count is defined as the inverse sum of its similarity to all other items within $S$, denoted as follows:
    \begin{align*}
    \text{count}(S_i) & = \frac{1}{\sum^K_{j=1} \text{Sim}(S_i, S_j)}, \\
    \text{Sim}(S_i, S_j) & = \cos{(\text{embed}(S_i), \text{embed}(S_j))},
    \end{align*}
    
    where $\text{embed}(\cdot)$ is parametrized by the \verb|parahprase-MiniLM-L6-v2| model provided by Sentence-Transformers. This is then used to measure the cardinality of the set, defined as the sum of its individual items:
    \begin{align*}
    \text{card}(S) & = \sum^K_{i=1}\text{count}(S_i).
    \end{align*}
    
    The heading soft recall (hsr) is then defined as:
    \begin{align*}
    \text{hsr} & = \frac{\text{card}(G \cap P)}{\text{card}(G)}, \\
    \text{card}(G \cap P) & = \text{card}(G) + \text{card}(P) - \text{card}(G \cup P).
    \end{align*}
    
    \item \textbf{Heading and Article Entity Recall:} These metrics indicate the proportion of named entities in the baseline gold review mentioned in the generated text, between headings or the full review texts, respectively. The implementation for this is also taken from \citet{shao_assisting_2024}. However, due to the presence of more scientific terms, we instead use the SciSpacy named entity tagger for all settings, specifically using the \verb|en_core_sci_lg| model, as it has the largest available vocabulary.
\end{itemize}

\section{LLM Evaluation}
\label{appendix:llme}

Here, we outline the details for LLM evaluation. For the specific LLM model, we used \verb|prometheus-7b-v2.0| using \cite{kim_prometheus_2024} with the same configurations outlined in \citet{shao_assisting_2024}, namely a temperature of 0.01, cumulative probability of top tokens of 0.95, maximum number of new tokens of 512, and repetition penalty of 1.03. Also, because of hardware limitations, the model had to be run with fp8-quantization and a memory usage value of 0.8. Note that no seeding was used here, with the justification being that it would encourage more diverse outputs from the model, thus better mimicking reviews made by separate experts. Moreover, to simulate the process of evaluation by Subject Matter Experts (SMEs) and account for randomness, the evaluation was performed three times for all articles, with the final score being the average between all runs. Furthermore, due to the model's context size, the evaluation for the criteria had to be done in the following ways:
\begin{itemize}
    \item For coverage and relevance, each generated review and its original article were split into their sections, with each section being evaluated separately using the same criteria. If any section is still too long for any of the two, then said section is split further into two parts each. The criteria for determining this is if the combined length of the generated and original section exceeds the context window (32,768 tokens) minus roughly three thousand tokens to account for the rubric and other components.
    \item For structure, evaluation was performed using only the outlines of both articles. This is because they are reasonably short and serve as a proper approximation for how the respective contents are arranged.
\end{itemize}

The rubric provided (shown in table \ref{table:criteria}) to the LLM contains scores ranging from 1 to 5, with the coverage, structure, and relevance evaluated for all generated literature reviews. Furthermore, the description for each score bracket is specified to accurately reflect the proper gradings. These were adapted from the rubric used by \citet{wang_autosurvey_2024}, with minor adjustments to accommodate for the output structure. Also, because three runs were performed per setting, the results reported are the means of the average scores and the averaged resulting standard deviations.

\section{Sample SME Evaluation}
\label{appendix:sme}

\begin{figure*}[t!]
    \centering
    \includegraphics[width=\textwidth]{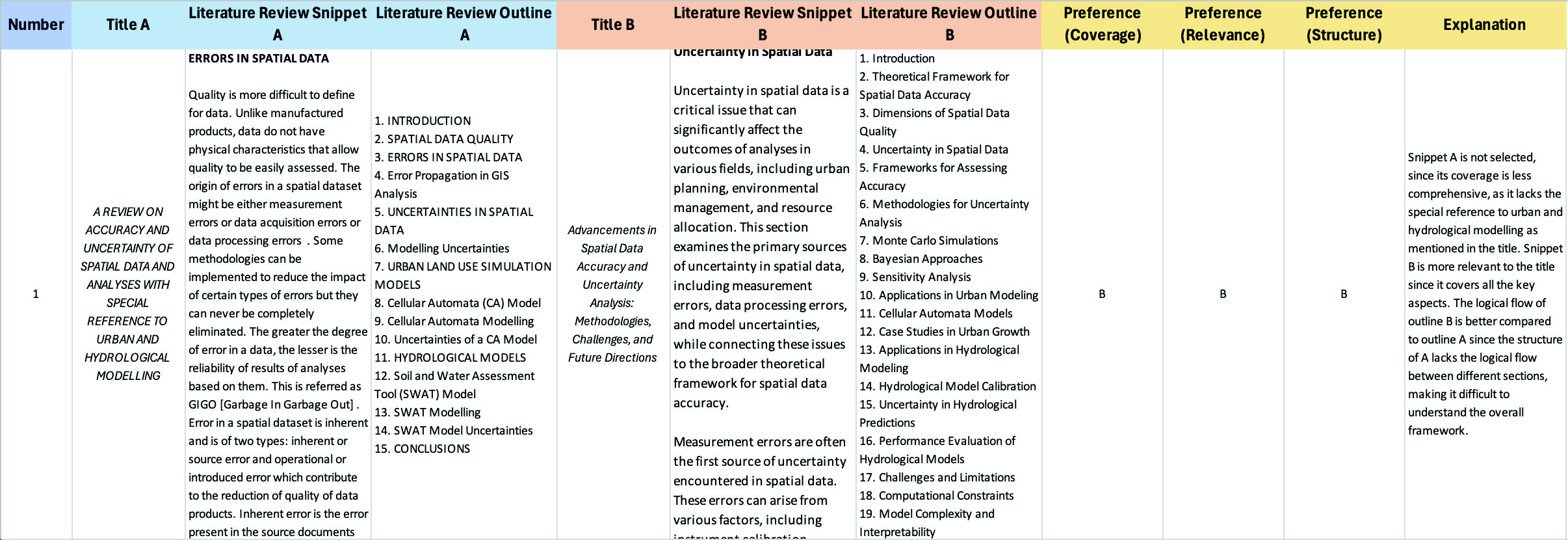}
    \caption{Annotation sample for the SciReviewGen dataset.}
    \label{appendix:sme_figure}
\end{figure*}

We include a sample annotation for the SciReviewGen dataset in figure \ref{appendix:sme_figure}. It can be seen that the SMEs were provided with the title, article snippet, and outline, which were then used to determine which article was qualitatively better.

\section{Citation Quality}
\label{appendix:citation_q}

This section provides an overview of how the citation quality F1-score functions. It uses an LLM (\verb|gpt-4o-mini|) to evaluate how well the article cites the given reference abstracts. More precisely, for a given article containing a set of claims $|C|$, an LLM performing Natural Language Inference denoted as $m$ determines the entailment of a claim $c_i$ given the references listed $Ref_i = \{r_{i_1}, r_{i_2}, \cdots\}$, returning 1 if the references support the claim and 0 otherwise.
        
This is then used to calculate the citation recall ($R_c$) and precision ($P_c$), which are defined as the following:
\begin{align*}
R_c & = \frac{\sum^{|C|}_{i=1} m(c_i, Ref_i)}{|C|}, \\
P_c & = \frac{\sum^{|C|}_{i=1} \sum^{|Ref_i|}_{j=1} m(c_i, Ref_i) \cap n(c_i, r_{i_j})}{\sum^{|C|}_{j=1}|Ref_i|},
\end{align*}
where $n(c_i, r_{i_j}) = (m(c_i, \{r_{i_j}\}) = 1) \cup (m(c_i, Ref_i \setminus \{r_{i_j}\}) = 0)$, which indicates if the paper $r_{i_j}$ is related to the claim $c_i$.

Given the consideration that longer articles are more difficult to cite properly for, a scaling factor was included for the recall, as it is reliant on the number of claims present. We argue that for a given article, the amount of effort required to properly cite references for the claims is roughly exponential due to the amount of information which has to be remembered from each paper and the relations between all papers. For example, if given references $Ref=\{r_1, r_2, r_3\}$, not only must one consider the relations between individual pairs ($(r_1, r_2), (r_2, r_3), (r_1, r_3)$), but also the potential relations between sets of papers. Thus, if a generated article is too short, its performance is penalized accordingly.

The scaling factor is defined as the following:
\begin{align*}
\text{Scaling} & = 1 - e^{(-k \times n_{\text{claims}})},
\end{align*}
where $k$ is a scaling factor (set as 0.01 based on the assumption that a paper has roughly 100 citations on average) and $n_{\text{claims}}$ is the average number of claims per article across the generated articles. By definition, this value asymptotically approaches 1 due to the exponential term when $n$ is large enough.
This is applied to the aggregate recall score to smooth out the weighting, which is then used to calculate a final F1-score.

For the references, only the abstracts were used, as doing so would replicate the process of how SMEs generally perform the same task. Furthermore, the implementation was taken from \citet{wang_autosurvey_2024} with some adjustments to better handle longer inputs. Specifically, assuming a claim-reference pair is too long, the reference is split into a few sections, and then the metric checks if any split supports the claim. This assumes the rest of the references are also supported by the sources as a result.

\section{Ablation and Modification}
\label{appendix:ablation}

\begin{figure}[t!]
    \centering
    \includegraphics[width=\linewidth]{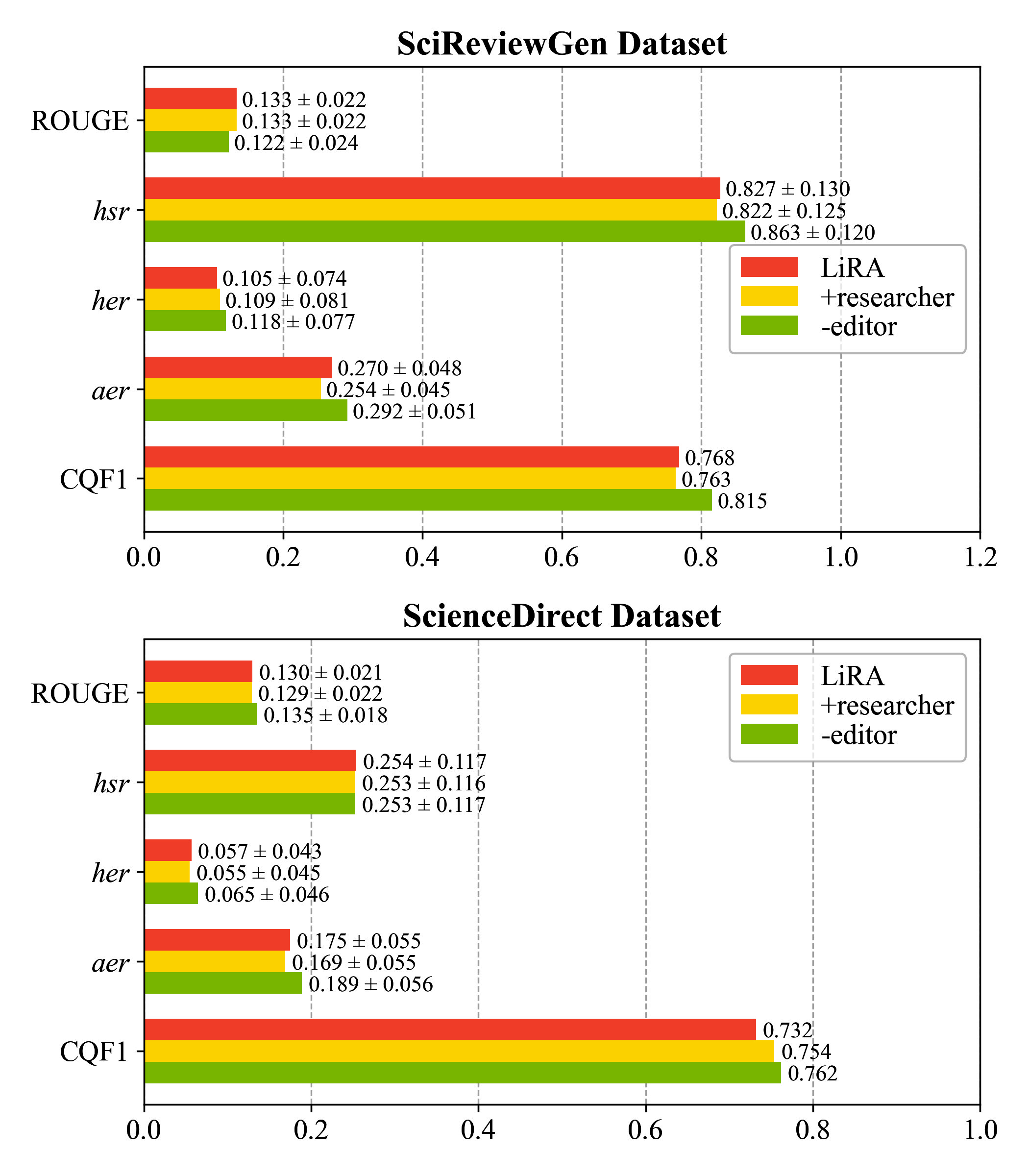}
    \caption{Textual similarity and citation quality results for the different reviewer model setting.}
    \label{appendix:sim_lira}
\end{figure}

\begin{figure}[t!]
    \centering
    \includegraphics[width=\linewidth]{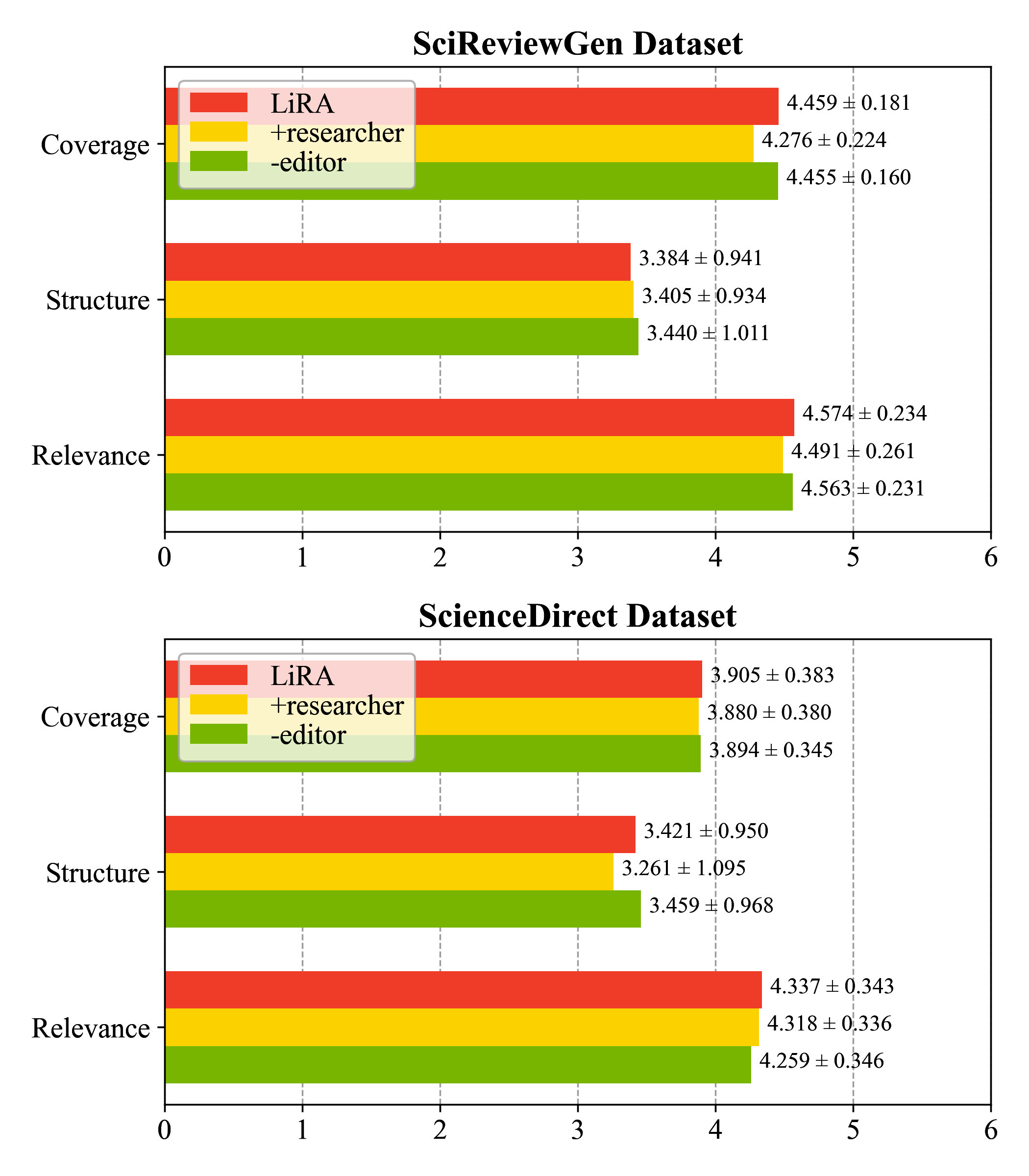}
    \caption{Prometheus evaluation results for the different reviewer model setting.}
    \label{appendix:prom_lira}
\end{figure}

Figures \ref{appendix:sim_lira} and \ref{appendix:prom_lira} show the results for additional studies relating to system design modification. More precisely, this involves two experiments, with the first being an ablation study of the editor model. This was performed because this component can be considered less impactful towards the output compared to most other agents in the system design.

The other experiment performed was the addition of a researcher agent. As a method of better imitating the process of literature review writing by humans, this agent takes each reference abstract and analyzes the contents based on several pre-determined questions based on existing literature \cite{snyder_literature_2019}. This results in a list of findings (i.e., key points, limitations) which are then utilized by the other agents in place of the original abstract.

The results indicate that the editor model has a generally small but negative impact towards the final results, though it does seem to improve the article coverage and relevance slightly in both cases. Meanwhile, the addition of a researcher agent also tends to decrease overall performance, again with marginal differences except in some metrics. As such, these results suggest that the editor agent in \textbf{LiRA} provides a trade-off in terms of article quality and similarity, while the researcher agent as it currently stands tends to reduce performance, likely because the abstracts are already full of information and do not require further processing.

\section{Retrieval Setting Statistical Tests}
\label{appendix:stats}

We include the statistical significance results for all results related to the retriever setting comparison in table \ref{appendix:stat_tests}. For all ROUGE- and recall-based metrics, the two-tailed Mann-Whitney U-test was used, as almost all of the results were not normally distributed (determined using the Shapiro-Wilk test on each set of results). Additionally, the results for the Prometheus gradings use the same test, though the results across all runs were averaged by sample first to combine the results first by group. 

All tests and normality checks were performed using the corresponding functions available in the \verb|scipy.stats| module. Moreover, the significance level was chosen to be $\alpha = 0.05$.

\begin{table}[t!]
\centering
\begin{tabular}{lcc}
\toprule
\textbf{Metric} & \textbf{\begin{tabular}[c]{@{}c@{}}U Statistic\\ value\end{tabular}} & \textbf{p-value}         \\ \midrule
ROUGE-L         & 8503.5                                                               & .226                     \\ 
\midrule
\textit{hsr}    & 8041.0                                                               & .690                     \\
\textit{her}    & 8414.5                                                               & .292                     \\
\textit{aer}    & 10097.5                                                              & \textbf{\textless{}.001} \\ 
\midrule
Coverage        & 8714.5                                                               & .114                     \\
Structure       & 7909.5                                                               & .864                     \\
Relevance       & 8970.5                                                               & \textbf{.042}            \\ 
\bottomrule
\end{tabular}
\caption{The statistical significance test results for the retriever experiment setting.}
\label{appendix:stat_tests}
\end{table}

\begin{table*}[t!]
\centering
\begin{tabular}{P{0.1\linewidth} P{0.05\linewidth} p{0.85\linewidth}}
\toprule
\textbf{Criteria}          & \multicolumn{1}{c}{\textbf{Score}} & \multicolumn{1}{c}{\textbf{Description}}     \\ 
\midrule
\multirow{10}{*}{Coverage}  & \multirow{2}{*}{5} & The literature review section comprehensively covers all key and peripheral topics, providing detailed discussions and extensive information.                                                                             \\
                           & \multirow{2}{*}{4}              & The literature review section covers most key areas of the topic comprehensively, with only very minor topics left out.                                                                                                   \\
                           & \multirow{2}{*}{3}              & The literature review section is generally comprehensive in coverage but still misses a few key points that are not fully discussed.                                                                                      \\
                           & \multirow{2}{*}{2}              & The literature review section covers some parts of the topic but has noticeable omissions, with significant areas either underrepresented or missing.                                                                     \\
                           & \multirow{2}{*}{1}              & The literature review section has very limited coverage, only touching on a small portion of the topic and lacking discussion on key areas.                                                                               \\ 
\midrule
\multirow{10}{*}{Structure} & \multirow{2}{*}{5} & The literature review outline is tightly structured and logically clear, with all sections and content arranged most reasonably, and transitions between adjacent sections smooth without redundancy.                    \\
                           & \multirow{2}{*}{4} & The literature review outline has good logical consistency, with content well arranged and natural transitions, only slightly rigid in a few parts.                                                                       \\
                           & \multirow{2}{*}{3} & The literature review outline has a generally reasonable logical structure, with most content arranged orderly, though some links and transitions could be improved such as repeated subsections.                         \\
                           & \multirow{2}{*}{2} & The literature review outline has weak logical flow with some content arranged in a disordered or unreasonable manner.                                                                                                    \\
                           & \multirow{2}{*}{1} & The literature review outline lacks logic, with no clear connections between sections, making it difficult to understand the overall framework.                                                                           \\ 
\midrule
\multirow{9}{*}{Relevance} & \multirow{2}{*}{5} & The literature review section is exceptionally focused and entirely on topic; the article is tightly centered on the subject, with every piece of information contributing to a comprehensive understanding of the topic. \\
                           & \multirow{2}{*}{4} & The literature review section is mostly on topic and focused; the narrative has a consistent relevance to the core subject with infrequent digressions.                                                                   \\
                           & \multirow{1}{*}{3} & The literature review section is generally on topic, despite a few unrelated details.                                                                                                                                     \\
                           & \multirow{2}{*}{2} & The literature review section is somewhat on topic but with several digressions; the core subject is evident but not consistently adhered to.                                                                             \\
                           & \multirow{2}{*}{1} & The literature review section is outdated or unrelated to the field it purports to review, offering no alignment with the topic.                                              \\ 
\bottomrule
\end{tabular}
\caption{The rubric criteria alongside their descriptions, adapted from \citep{wang_autosurvey_2024}.}
\label{table:criteria}
\end{table*}

\end{document}